%% file: ITSCvr23.tex
\documentclass[conference]{ieeeconf}
\IEEEoverridecommandlockouts
\usepackage{cite}
\usepackage{amsmath,amssymb,amsfonts}
\usepackage{algorithmic}
\usepackage{lipsum}
\usepackage{multirow}
\usepackage{graphicx}
\usepackage[caption=false]{subfig}
\usepackage{textcomp}
\usepackage{xcolor}
\def\BibTeX{{\rm B\kern-.05em{\sc i\kern-.025em b}\kern-.08em
    T\kern-.1667em\lower.7ex\hbox{E}\kern-.125emX}}
\begin{document}

\title{\LARGE \bf Digital twin in virtual reality for human-vehicle interactions in the context of autonomous driving}



\author{S. Martín Serrano$^{1}$, R. Izquierdo$^{1}$,  I. García Daza$^{1}$, M. A. Sotelo$^1$, D. Fern\'andez Llorca$^{1,2}$
\thanks{$^{1}$Computer Engineering Department, Universidad de Alcal\'a, Alcal\'a de Henares, Spain.
        {\tt\small sergio.martin@uah.es}}%
        \newline
\thanks{$^{2}$European Commission, Joint Research Centre, Seville, Spain.}
}

\maketitle

\input{00_abstract.tex}
\input{01_introduction.tex}
\input{02_experiment_configuration.tex}
\input{03_questionnaire_and_variables.tex}
\input{04_results.tex}

\input{05_conclusions_and_FW.tex}

\section*{Acknowledgment}
This work was funded by Research Grants PID2020-114924RB-I00 and PDC2021-121324-I00 (Spanish Ministry of Science and Innovation) and partially by S2018/EMT-4362 SEGVAUTO 4.0-CM (Community of Madrid). D.F.LL. acknowledges funding from the HUMAINT project by the Joint Research Centre of the European Commission.

\section*{Disclaimer}
The views expressed in this article are purely those of the authors and may not, under any circumstances, be regarded as 
an official position of the European Commission. 

\bibliographystyle{IEEEtran}
\bibliography{ITSCvr23}

\end{document}

%% file: 00_abstract.tex
\begin{abstract}
This paper presents the results of tests of interactions between real humans and simulated vehicles in a virtual scenario. Human activity is inserted into the virtual world via a virtual reality interface for pedestrians. The autonomous vehicle is equipped with a virtual Human-Machine interface (HMI) and drives through the digital twin of a real crosswalk. The HMI was combined with gentle and aggressive braking maneuvers when the pedestrian intended to cross. The results of the interactions were obtained through questionnaires and measurable variables such as the distance to the vehicle when the pedestrian initiated the crossing action. The questionnaires show that pedestrians feel safer whenever HMI is activated and that varying the braking maneuver does not influence their perception of danger as much, while the measurable variables show that both HMI activation and the gentle braking maneuver cause the pedestrian to cross earlier.
\end{abstract}


%% file: 01_introduction.tex
\section{Introduction}
Autonomous driving technology has emerged as a promising solution for enhancing road safety, reducing traffic congestion, and improving the overall driving experience. As the deployment of autonomous vehicles becomes more widespread, understanding the interactions between humans and these vehicles becomes increasingly critical \cite{Dey2018, Llorca2023}. 

Previous research has focused on employing diverse strategies to effectively communicate the current state or intention of autonomous vehicles to vulnerable road users (VRUs), such as pedestrians or cyclists \cite{Mirnig2018}. Relying on the information provided can help users make better decisions, for example, when crossing the road. The design of external Human Machine Interfaces (eHMIs) that enable communication can include external displays, light or sound signals \cite{Dey2020,Ranasinghe2020}.

Pedestrians evaluate multiple factors to determine the safety of road crossing, such as vehicle speeds and movements, safety gap sizes or location familiarity. While implicit interaction, such as perceived speed or gap size, may be the primary basis for crossing decisions, as suggested by some studies \cite{Clamann2017, Dey2017, Zimmermann2017}, some experiments indicate that explicit communication may also play a role, depending on the distance between the pedestrian and the vehicle \cite{Dey2019}.

Conversely, the effectiveness of virtual reality (VR) in conducting human factors research within the realm of interface design has been widely acknowledged, due to issues of safety and control of the experimental setting for each trial \cite{Nascimento2019}. A pedestrian VR simulator can generate the possibility of participants being hit by the vehicle and create the real sense of fear in them \cite{Deb2018}. The ability of the simulator to make pedestrians exhibit very realistic behavior (i.e., to minimize the \emph{behavioral gap} with respect to a real situation) justifies the use of VR in interaction research. In addition, VR platform provides an efficient method for measuring pedestrian travel behavior via headset tracking data and synchronized video recordings \cite{Sween2016}.

\begin{figure}
\centerline{\includegraphics[width=0.92\columnwidth]{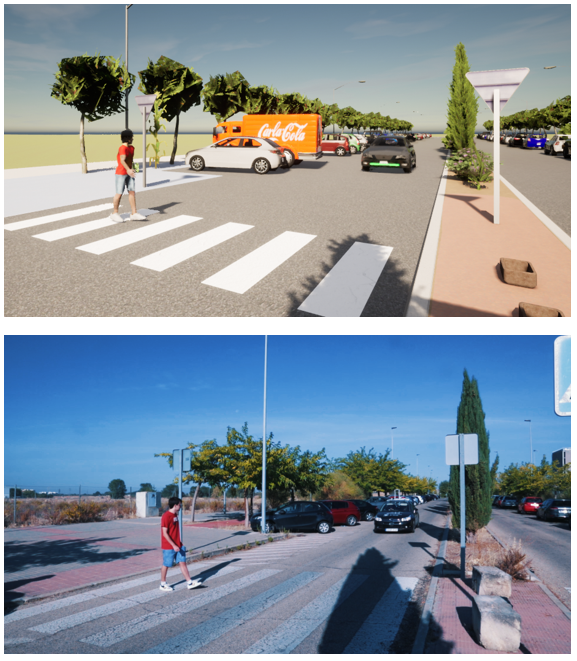}}
\caption{Digital twin for human-vehicle interaction in autonomous driving. Virtual (above) and real (bottom) setting.}
\label{fig:schematic1}
\end{figure}

In our previous work \cite{CarlaCHIRA}, \cite{CarlaCHIRA23}, we presented a framework to enable real-time interaction between pedestrians and CARLA simulator using immersive virtual reality and human motion capture systems. In the following, we present the application of this framework to develop a digital twin of a real scenario, and replicate the field experiments carried out in real-world driving conditions \cite{Izquierdo2023} (see Fig. \ref{fig:schematic1}). The experiments are focused in studying human-vehicle interaction in crosswalks through the use of external HMIs and implicit communication based on the motion of the vehicle. We evaluate the effectiveness of different forms of communication between the autonomous vehicle and real pedestrians immersed in a virtual crosswalk scenario, which will then be used in a future work to establish a comparison with the same study at a real twin crosswalk and provide a reliable measure of the \emph{behavioral gap}.

\section{Related work}
VR environments are frequently used to investigate pedestrian crossing actions, providing a safer alternative to real-world experiments while also enabling greater flexibility and replicability in constructing experimental scenarios \cite{Nascimento2019}. We summarize the works that have employed VR to simulate an immersive virtual pedestrian crossing experience.

Most of the approaches focus on studying the communication between autonomous vehicles and pedestrians based on the design of different eHMI systems, such as eyes added to the car that look at the pedestrian to indicate their intention to stop \cite{Chang2017}, or a human-like visual embodiment in the driver's seat \cite{Furuya2021}. \cite{Shuchisnigdha2020-2} shows greater acceptance of written messages (e.g., "walk") than conveyed by images such as a walking silhouette or a raised hand. Malfunctioning external displays of the car lead pedestrians to ignore it and aggravate the effects of overtrust in the autonomous vehicle, so the design of external communication must avoid misleading information \cite{Hollander2019}, \cite{Shuchisnigdha2020}. Certain studies explore ways in which pedestrians can communicate their intentions to autonomous vehicles (e.g. through gestures \cite{Gruenefeld2019}) while the vehicle displays its locomotion and reacts to the pedestrian's language. A novel application of eHMI incorporates vehicle directional information in scenarios where a pedestrian is in close proximity to a collision and increases pedestrians' self-reported understanding of the car's intent \cite{Bazilinskyy2022}.

To our knowledge, all previous work has used VR scenarios developed with Unity, and has not operated in a simulator specifically dedicated to autonomous driving. In this paper, the chosen proposal enables a VR environment in the open-source CARLA simulator with an interactive pedestrian interface \cite{CarlaCHIRA}. In addition, experiments are conducted on the digital twin of a pre-existing crosswalk with a virtual replica of an autonomous vehicle, exploiting all the advantages of working with digital twins and virtual reality \cite{Hoffmann2022, Jagannath2022}. Another remarkable aspect is that body movements are not only recorded on video, but also registered with a motion capture system to collect accurate data on pedestrian responses to the variables under study. 

%% file: 02_experiment_configuration.tex
\section{Experiment Description}
The designed experiment aims to produce an interaction  between a simulated autonomous vehicle and a real pedestrian within a virtual environment. Since the traffic scenario is generated by the CARLA autonomous driving simulator \cite{carla2017}, we are not only able to conduct the experiment under controlled conditions, but also to evaluate the result by analysing the participants performance through direct and indirect measurements and questions. This section describes the scene recreated in the simulator, the immersive and maneuverable virtual reality interface between the real subject and the virtual world, and the different experiment settings.

\subsection{Experiment Configuration}
As the objective of this experiment is to carry out a study of communication interfaces for autonomous vehicles under simulated conditions but whose result can be compared with the one obtained in the real setting, we have developed a digital twin of a real crosswalk from a georeferenced map to establish the same dimensions of the road, including the same arrangement of the rest of elements (i.e., traffic signs, other parked vehicles, vegetation) to faithfully reproduce the visibility conditions.

As shown in Figure \ref{fig:schematic1}, the vehicle circulates autonomously on the road when it reaches the crosswalk just as the pedestrian intends to cross perpendicular to the opposite sidewalk. The pedestrian can detect the vehicle a few meters ahead of the crosswalk before starting the crossing action, and lighting and weather conditions are favorable. The participant is instructed to wait with their back to the roadway until the vehicle is 40 meters from the crosswalk and is then instructed to turn around and move towards the roadway. The vehicle speed is 30 km/h and it starts a braking maneuver using a constant deceleration until it comes to a complete stop at the edge of the crosswalk to yield to the pedestrian.

\begin{figure}
\centering
\subfloat[]{\includegraphics[width=0.49\columnwidth]{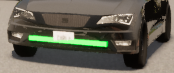}%
\label{fig:grail_green}}
\hfil
\subfloat[]{\includegraphics[width=0.49\columnwidth]{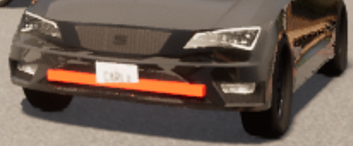}%
\label{fig:grail_red}}
\caption{External HMI activated communicating a green status (a) and a red status (b).}
\label{fig:external_HMI}
\end{figure}

\subsection{Test Variations}
The virtual autonomous vehicle is equipped with an external communication interface, a so-called GRAIL (Green Assistant Interfacing Light) \cite{GRAIL}, which is represented in the simulator by a bar along the
entire front of the car that changes color to communicate its intentions and current status to other agents on the road. As Figures \ref{fig:grail_green} and \ref{fig:grail_red} show, the interface emits a red color to warn the pedestrian that the vehicle has not detected any obstacles in its path and that it does not plan to execute any braking maneuver, while the green color anticipates a stop to avoid a collision. It is also possible that the interface is turned off so the pedestrian does not have any information about the vehicle status. 

Furthermore, we add implicit form of communication by varying the braking profile. To study whether the pedestrian perceives to be in a situation of greater risk depending on the braking profile, we define a gentle braking maneuver, in which the vehicle decelerates at -0.9 m/s\textsuperscript{2}, and a second, more aggressive braking maneuver, when the vehicle decelerates at -1.8 m/s\textsuperscript{2}. In both cases, the vehicle reduces its 30 km/h speed to a complete stop, but the aggressive maneuver simulates less vehicle anticipation of the encounter.

\subsection{Test Batch}
We designed five tests to assess the influence of each communication technique over the pedestrian level of confidence and their perceived level of safety during the experiment. Table \ref{tab:testsetup} shows the variations in the braking maneuver as well as activation of the external communication interface. 
Test number 0 purpose is to prime the participants for the environment and potential risk of a collision. All tests were performed in random order except test number 0, which was always performed first for each participant.

\begin{table}[htbp]
\renewcommand{\arraystretch}{1.1}
\caption{Experimentation Tests settings}
\begin{center}
\begin{tabular}{c|c|c|c|c}
\textbf{Test}& \textbf{Braking} & \textbf{External} & \textbf{Stop}\\
\textbf{Number} & \textbf{Maneuver} & \textbf{HMI} &  \\
\hline
0   & -             & -     & No \\
1   & Gentle        & -     & Yes\\
2   & Aggressive    & -     & Yes\\
3   & Gentle        & GRAIL & Yes\\
4   & Aggressive    & GRAIL & Yes\\
\end{tabular}
\label{tab:testsetup}
\end{center}
\end{table}

\subsection{Virtual Reality Setup}
In order to allow the experiment to be conducted within a virtual environment, we harness the full immersive system for pedestrians described in \cite{CarlaCHIRA} which adds some features to the CARLA simulator such as real-time avatar control, positional sound and the body tracking of the subject interacting with the scene through virtual reality. In this way, we take advantage of all the different options that CARLA offers to simulate specific traffic scenarios while there is a real subject playing the role of a pedestrian and being part of the simulation. We use Oculus Quest 2, created by Meta, as a head mounted device (HMD) and Perception Neuron Studio (PNS) motion capture system for full-body tracking \cite{PNS2022}. Quest 2 is connected to PC via WiFi and projects onto their lenses the CARLA spectator view. At the same time, the captured pose and motion of the subject is integrated into the virtual scenario, so the simulated sensors attached to the autonomous vehicle (i.e., radar, LiDAR, cameras) can be aware of their presence. So that the development of the experiment was not hindered, we reserved a preset area (3 x 8 meters) free of obstacles where the participant could act as a real pedestrian inside the simulation.

\subsection{Participants}
18 volunteers from inside and outside the University area, consisted of 12 male and 6 female who ranged in age from 24 to 62, agreed to participate in the experiment. Most of them had never had any virtual reality experience before and they were informed about the risk of dizziness or disorientation. Fortunately, all the participants felt good during the experiment and there were no incidents.

%% file: 03_questionnaire_and_variables.tex
\section{Experiment Evaluation}
This section presents the tools used to evaluate the influence of the communication interfaces over the confidence level of the participants. The tests include an explicit and an implicit communication interface, the external HMI and the braking maneuver profile respectively. For this analysis, after each test in Table \ref{tab:testsetup}, they were asked to fill out a questionnaire about their subjective perception of the interaction. In addition, direct measurement variables were registered from the scenario to evaluate changes in their observable behavior.

\begin{figure}
\centerline{\includegraphics[width=\columnwidth]{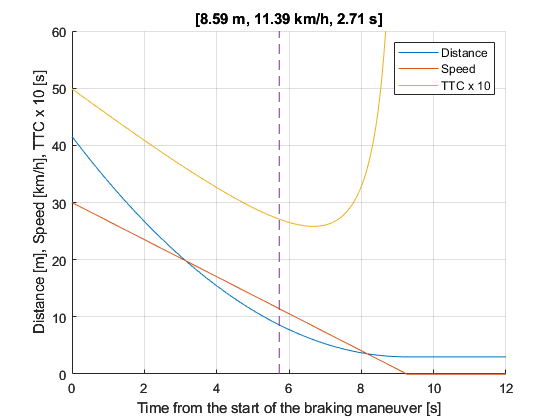}}
\caption{Crossing event example in the gentle braking maneuver.}
\label{fig:varibles_at_crossing_gentle}
\end{figure}

\subsection{Questionnaire}
Throughout the experiment, the participants had short rest periods between tests in which they did not leave the virtual reality when a researcher asked them the following questions about their last interaction with the vehicle:

\begin{itemize}
\item[--] Q1: \emph{What was your level of confidence that the vehicle would stop and yield to you?}
\item[--] Q2: \emph{How did you perceive the braking of the vehicle?}
\item[--] Q3: \emph{Has the visual communication interface improved your confidence to cross?}
\end{itemize}

Answers to these questions are tabulated on a 7-step Likert scale \cite{joshi2015likert} and allow to study the influence of communication interfaces from the subjective point of view of the pedestrian. 

\subsection{Direct Measurements}
In addition to having specific control over traffic conditions, CARLA simulations enable access to all agents and environment variables so we directly obtain the participant's location on the map and their full-body pose. The reconstruction of their trajectory allows us to generate synthetic sequences from multiple points of view based on their real behavior and to extract some valuable parameters such as the crossing decision event, the crossing event or their eye contact with the vehicle. The experiments are recorded and can be replayed to compare results for different sensors or configurations.
The direct measurements used to quantitatively analyze the interaction during the experimentation are the distance to the pedestrian, the vehicle speed and the time-to-collision (TTC) computed as  $TTC=d/v$. Figure \ref{fig:varibles_at_crossing_gentle} shows the evolution of these variables in the gentle braking maneuver, as well as a crossing event example.

\subsection{Crossing Event}
\begin{figure}[t]
  \centering
  \includegraphics[width=1\linewidth]{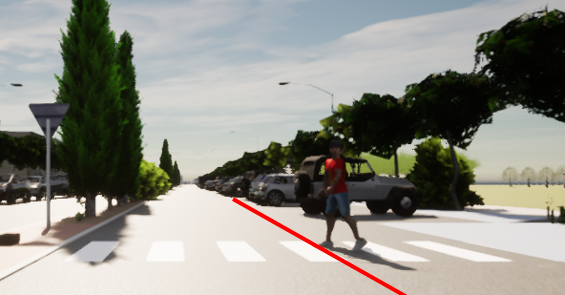}
  \caption{The crossing event is defined when the pedestrian enters the vehicle lane and is exposed to a possible collision.}
  \label{fig:vehicle_lane}
\end{figure}

The crossing event is defined as the pedestrian entering the vehicle lane an exposing themselves to a possible collision, as shown in Figure \ref{fig:vehicle_lane}. It is considered a metric to evaluate the behaviour of the pedestrian: if crosses earlier, s/he feels more confident in the vehicle. The crossing event is used instead of the previous crossing decision event (i.e., the moment when the pedestrian makes the decision to cross the road, which is more difficult to measure) because it can be unequivocally identified  by means of the lane marking. If the subject perceives that the situation is more risky and hesitates to cross, we detect this in a subsequent crossing event. Both braking maneuvers are repeated in every experiment, so the crossing event is needed for directly observable measurements to be meaningful in the study.  

%% file: 04_results.tex
\section{Results}
\begin{figure*}
\centering
\subfloat[]{\includegraphics[width=0.195\textwidth]{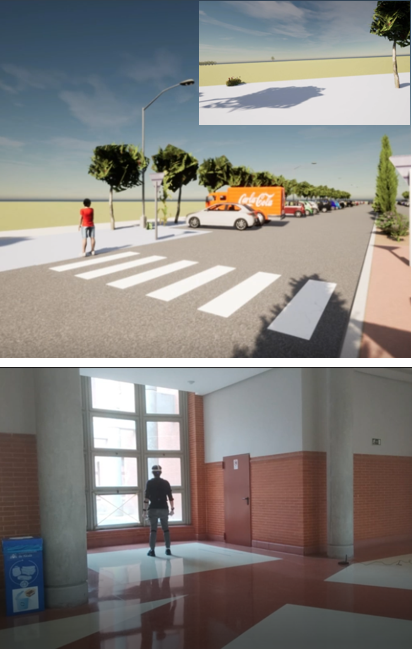}%
\label{fig_ped_1}}
\hfil
\subfloat[]{\includegraphics[width=0.195\textwidth]{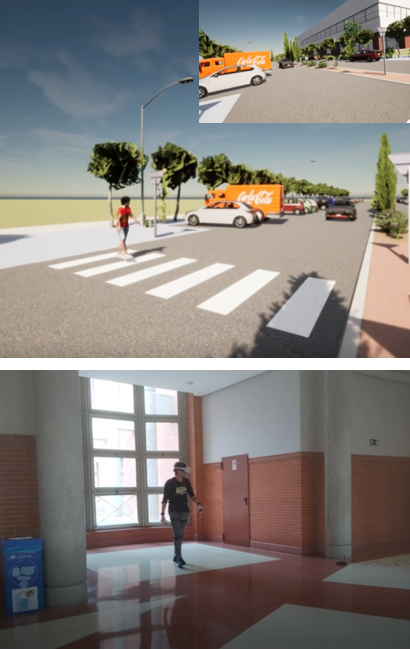}%
\label{fig_ped_2}}
\hfil
\subfloat[]{\includegraphics[width=0.195\textwidth]{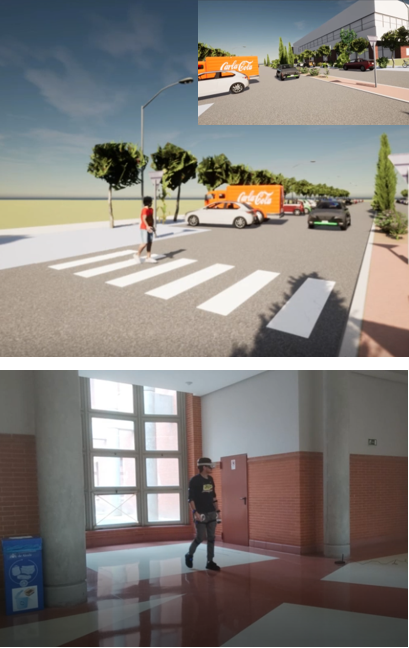}%
\label{fig_ped_3}}
\hfil
\subfloat[]{\includegraphics[width=0.195\textwidth]{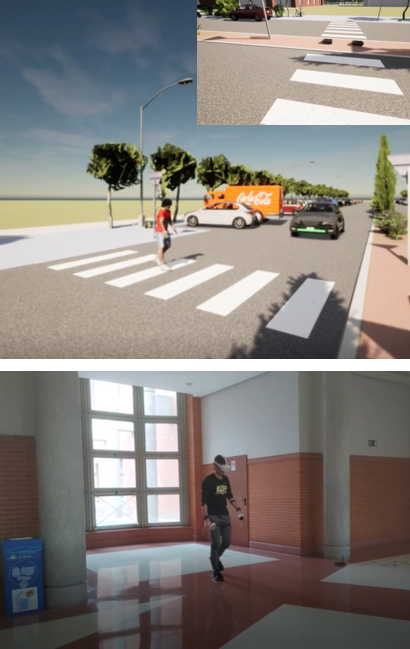}%
\label{fig_ped_4}}
\hfil
\subfloat[]{\includegraphics[width=0.195\textwidth]{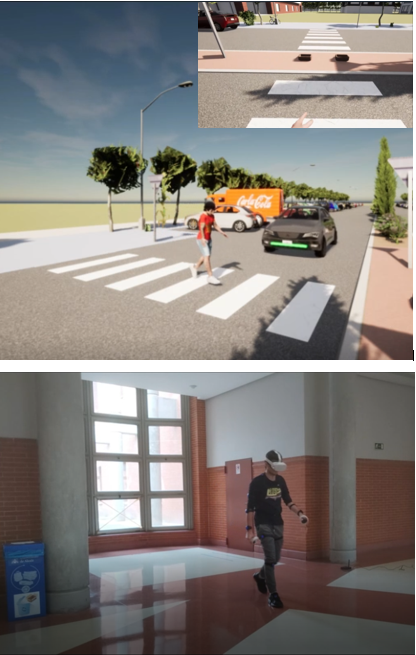}%
\label{fig_ped_5}}
\caption{Interaction example between pedestrian and virtual vehicle equipped with external HMI: (a) The pedestrian starts with their back to the crosswalk and is told to turn around when the vehicle approaches. (b) The pedestrian makes eye contact with the vehicle and hesitate to cross. (c) The external HMI switches from red to green. (d) The pedestrian enters the vehicle lane establishing the crossing event. (e) The pedestrian crosses the road.}
\label{fig:example_interacion_exterior}
\end{figure*}

This section presents the results obtained both in the questionnaires and in the labeling of the direct measurements. We search for significant differences between the tests from Table \ref{tab:testsetup} to determinate the utility of the communication interfaces involved. In Figure \ref{fig:example_interacion_exterior} we can observe an interaction example with the external HMI activated. The virtual reality headset projects the crosswalk onto its lenses and allows displacement through the scenario. To compare the responses to the questionnaires and the direct measurements we use the Wilcoxon signed-rank test and the Student t-test respectively.

\begin{figure}
\centerline{\includegraphics[width=\columnwidth]{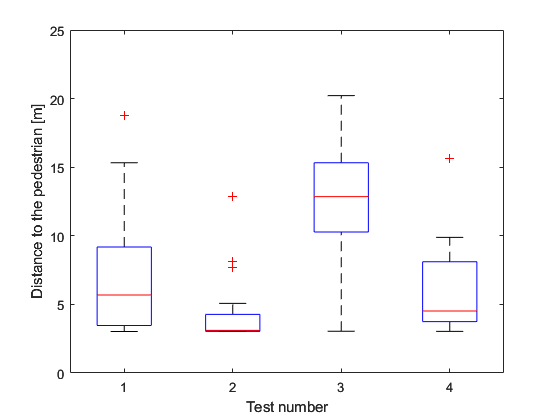}}
\caption{Box-plots of the distances to the pedestrian at the crossing event.}
\label{fig:boxplot_distance}
\end{figure}

\subsection{Questionnaire Results}
The Wilcoxon Signed Rank test \cite{woolson2007wilcoxon} is a non-parametric statistical hypothesis test used to determinate whether the difference between two related samples taken from the same population is statistically significant. The alternative hypothesis matrix showed in Table \ref{tab:wilcoxon} expresses categorical statements which compare the answers to the questions from Q1 to Q3 obtained in each test from the experiment. A check-mark in a specific cell means the null hypothesis $H_0:\mu_i\leq\mu_j$ is rejected and  the alternative hypothesis $H_1:\mu_i>\mu_j$ is accepted when comparing the answers provided in test $i$ (left column) and in test $j$ (top row). Rejecting $H_0$ and accepting $H_1$ implies there is a significant difference in the answers and that the score in test $i$ is higher than in test $j$.

\begin{table}[htbp]
\renewcommand{\arraystretch}{1.1}
\caption{Wilcoxon Signed Rank test, Q1-Q3, $\alpha$=0.05}
\begin{center}
\begin{tabular}{ccc|p{1cm}p{1cm}p{1cm}p{1cm}}
\multicolumn{3}{c|}{\textbf{$H_1:\mu_i>\mu_j$}}& \multicolumn{4}{c}{\textbf{Test number $j$}} \\
 \multicolumn{3}{c|}{} & 1 & 2 & 3 & 4\\
\hline
\multirow{12}{*}{\rotatebox[origin=c]{90}{\textbf{Test number $i$}}} & \multirow{4}{*}{\rotatebox[origin=c]{90}{\textbf{Q1}}}     & 1   & --   &    &      &\\
                                            &&2   &    & --    &     & \\
                                            &&3   & \checkmark    & \checkmark     & --    & \checkmark\\
                                            &&4   & \checkmark    & \checkmark     &     & --\\

\cline{3-7}
&\multirow{4}{*}{\rotatebox[origin=c]{90}{\textbf{Q2}}} & 1   & --   &     &     &\\
                                                        &&2   & \checkmark    & --    & \checkmark     & \checkmark\\
                                                        &&3   &    &     & --    &\\
                                                        &&4   & \checkmark    &     & \checkmark     & --    \\

\cline{3-7}
&\multirow{4}{*}{\rotatebox[origin=c]{90}{\textbf{Q3}}} & 1   & --   &     &     &\\
                                                        &&2   &    & --    &     &     \\
                                                        &&3   & \checkmark    & \checkmark     & --    &     \\
                                                        &&4   & \checkmark    & \checkmark     &     & --    \\

\end{tabular}
\label{tab:wilcoxon}
\end{center}
\end{table}

Based on the results showed on Table \ref{tab:wilcoxon}, we cannot state that the gentle braking maneuver with the external HMI non-activated contributes to increase the pedestrian's confidence in the vehicle (Q1: test1 vs test2), but we can do state that the gentle braking maneuver with the external HMI activated does contribute to increase  the pedestrian's confidence in the vehicle (Q1: test3 vs test4). The external HMI does contribute to increase the pedestrian's confidence in the vehicle (Q1: test3 vs test1 and test4 vs test2) and pedestrians perceived the aggressive braking maneuver as “more aggressive” or “less conservative” than the gentle braking maneuvers (Q2: test2 vs test1 and test4 vs test3).

\subsection{Direct Measurements Results}
In the direct measurements analysis we use the Student's t-test \cite{student} that determines if there is a significant difference between the means of two samples groups. The alternative hypothesis matrix is represented in Table \ref{tab:student}. A check-mark in a specific cell means the null hypothesis $H_0:\mu_i\leq\mu_j$ is rejected and  the alternative hypothesis $H_1:\mu_i>\mu_j$ is accepted when comparing the direct measurements labeled on test $i$ (left column) and test $j$ (top row). Rejecting $H_0$ and accepting $H_1$ implies distance, speed and/or TTC at the crossing event in test $i$ are significantly higher than in test $j$. Figure \ref{fig:boxplot_distance} shows the box-plots of the distance between the pedestrian and the vehicle in the labeled crossing event in each trial.

\begin{table}[htbp]
\renewcommand{\arraystretch}{1.1}
\caption{Student t-test, $\alpha$=0.05}
\begin{center}
\begin{tabular}{ccc|p{1cm}p{1cm}p{1cm}p{1cm}}
\multicolumn{3}{c|}{\textbf{$H_1:\mu_i>\mu_j$}}& \multicolumn{4}{c}{\textbf{Test number $j$}} \\
 \multicolumn{3}{c|}{} & 1 & 2 & 3 & 4\\
\hline
\multirow{12}{*}{\rotatebox[origin=c]{90}{\textbf{Test number $i$}}} & \multirow{4}{*}{\rotatebox[origin=c]{90}{\textbf{Distance}}}   & 1   & --   & \checkmark    &      &\\
                                                &&2   &    & --    &     &\\
                                                &&3   & \checkmark    & \checkmark     & --    & \checkmark\\
                                                &&4   &    & \checkmark     &     & --\\
\cline{3-7}
&\multirow{4}{*}{\rotatebox[origin=c]{90}{\textbf{Speed}}}  & 1   & --   & \checkmark     &     &\\
                                                            &&2   &    & --    &     &     \\
                                                            &&3   & \checkmark    & \checkmark     & --    & \checkmark     \\
                                                            &&4   &    & \checkmark     &     & --    \\
\cline{3-7}
&\multirow{4}{*}{\rotatebox[origin=c]{90}{\textbf{TTC}}}    & 1   & --   &     &     &\\
                                                            &&2   & \checkmark    & --    & \checkmark     & \checkmark     \\
                                                            &&3   &    &     & --    &     \\
                                                            &&4   &    &     &     & --    \\
\end{tabular}
\label{tab:student}
\end{center}
\end{table}

Based on the results showed on Table \ref{tab:student} we can state that the gentle braking maneuver contributes to an increase in distance at the crossing event (distance: test1 vs test2 and test3 vs test4) and the external HMI does contribute to increase the distance at the crossing event (distance: test3 vs test1 and test4 vs test2). The alternative hypothesis matrix of the vehicle speed confirms the previous statements: the greater the distance to the pedestrian, the greater the vehicle speed due to its constant deceleration. The aggressive braking maneuver with the external HMI non-activated increases the time-to-collision (TTC: test2 vs test1, test2 vs test3 and test2 vs test4).

\subsection{Results Discussion}
In the responses to Q1, participants express greater confidence whenever the external HMI is activated. It should be noted that the virtual environment does not distort the appreciation of the braking maneuver, since in the responses to Q2 the aggressive maneuver is always described as “more aggressive” than the gentle maneuver. However, it draws our attention that the non-activation of the external HMI in combination with the aggressive maneuver implies that the same braking maneuver is perceived as even more aggressive (Q2: test2 vs test4). We can make the statement that the activation of the external HMI has much more influence on the risk perception of the participant, than the type of maneuver used in the test.

If we look at the distance to the pedestrian and the vehicle speed in Table \ref{tab:student}, we obtain the same information of the crossing event since the braking maneuver follows a constant deceleration. If participants cross earlier, we can infer they feel more confident, because the vehicle is farther away from coming to a complete stop. Despite the fact that in the questionnaire the participants claimed that they mostly felt safer with the external HMI activated, even with the aggressive maneuver (Q1: test4 vs test1), in practice they also crossed the road earlier when the vehicle followed a gentle braking maneuver. In any case, the activation of the external HMI continues to have a very high influence in making the decision to cross sooner. The non-activation of the external HMI in combination with the aggressive braking maneuver sharply increases the time-to-collision. We suggest that this is because the participants perceive the situation as high risk and wait for the vehicle to reduce its speed to almost zero.

%% file: 05_conclusions_and_FW.tex
\section{Conclusions and Future Work}
Both the questionnaire and direct measurements support the virtual external HMI increases the pedestrian confidence and also leads to an earlier crossing event. On the other hand, in the questionnaire participants do not express greater confidence in gentle braking maneuver compared to aggressive braking maneuver without the activation of the external HMI. We suggest that participants state to feel safer only by the activation of the external HMI due to its high visibility in the virtual scenario, although the gentle braking maneuver also entails an earlier crossing event. 

As future work, it is intended to compare results on the acceptance of implicit and explicit communication interfaces in the real setting (see Fig. \ref{fig:schematic1}). The final goal is to verify that the difference in pedestrian responses in the virtual and real scenario are not statistically significant, thus validating the method to include realistic behaviors in the simulations through the proposal of a fully immersive VR system for pedestrians in the CARLA simulator \cite{CarlaCHIRA}.